\documentclass[10pt,twocolumn,letterpaper]{article}

\usepackage{iccv}      %

\definecolor{iccvblue}{rgb}{0.21,0.49,0.74}
\usepackage[pagebackref,breaklinks,colorlinks,allcolors=iccvblue]{hyperref}

\usepackage{xr}
\usepackage{amsmath}
\usepackage{microtype}
\usepackage{graphicx}
\usepackage{url}
\usepackage{physics}
\usepackage{booktabs} %
\usepackage{multirow}
\usepackage{colortbl}
\usepackage{rotating}
\usepackage{enumitem}
\usepackage{lipsum}
\usepackage{subcaption}
\usepackage{xspace}
\usepackage{algorithm}
\usepackage{algpseudocode}
\usepackage{amssymb}
\usepackage{pifont}

\usepackage{makecell}

\usepackage{amsmath,amsfonts,bm}

\def\eqref#1{equation~\ref{#1}}

\def\1{\bm{1}}

\DeclareMathAlphabet{\mathsfit}{\encodingdefault}{\sfdefault}{m}{sl}
\SetMathAlphabet{\mathsfit}{bold}{\encodingdefault}{\sfdefault}{bx}{n}

\DeclareMathOperator*{\argmin}{arg\,min}

\def\paperID{5939} %
\def\confName{ICCV}
\def\confYear{2025}

\usepackage{xcolor}

\definecolor{cA}{HTML}{0D68AB}
\definecolor{cC}{HTML}{0B4EA3}
\definecolor{cE}{HTML}{08349A}

\newcommand{\CA}[1]{\textcolor{cA}{#1}}
\newcommand{\CB}[1]{\textcolor{cC}{#1}}
\newcommand{\CC}[1]{\textcolor{cE}{#1}}

\newcommand{\modelname}{TesserAct\xspace}
\title{
\raisebox{-8pt}{\includegraphics[width=2em]{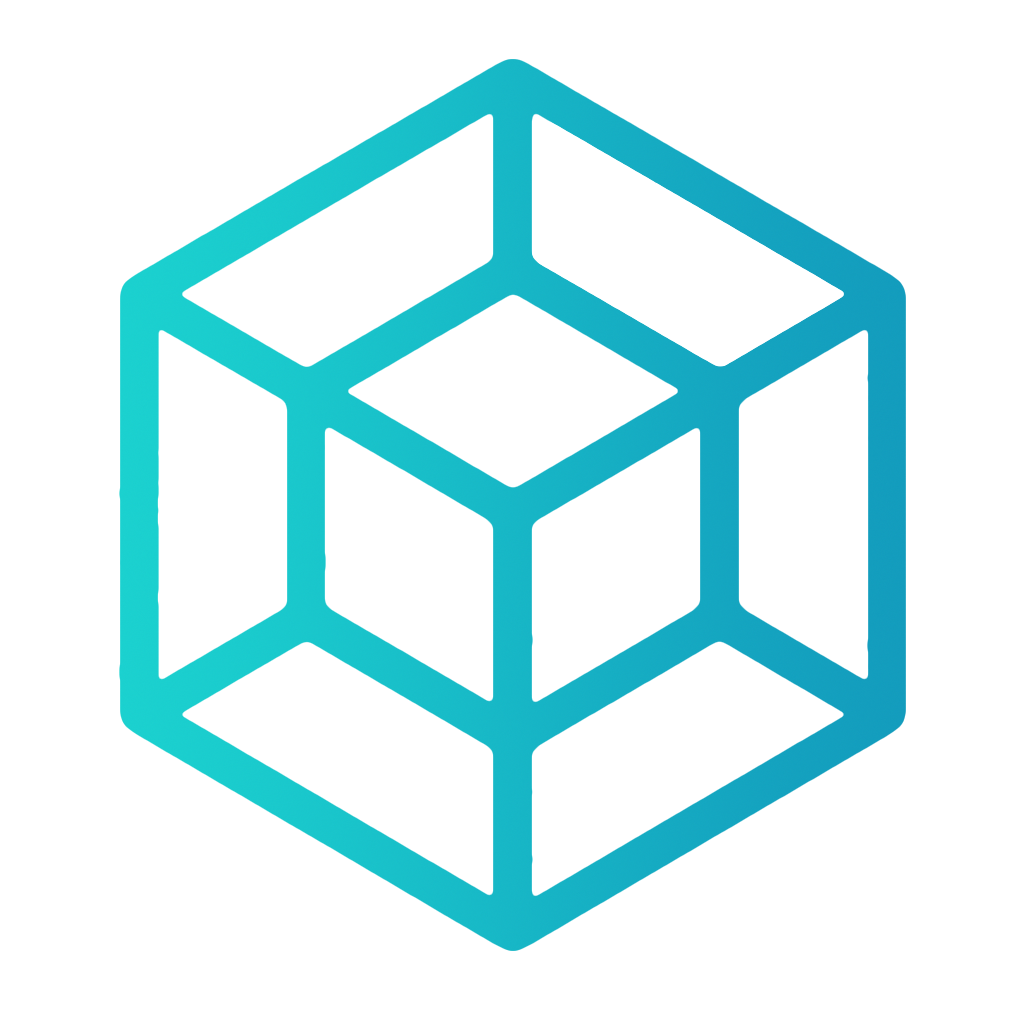}}
\CA{Tes}\CB{ser}\CC{Act}: Learning 4D Embodied World Models
\vspace{-5mm}
}

\author{
Haoyu Zhen$^{1\ *}$ \quad
Qiao Sun$^{1\ *}$ \quad
Hongxin Zhang$^{1}$ \quad
Junyan Li$^{1}$ \quad
Siyuan Zhou$^{2}$ \quad \\
Yilun Du$^{3}$ \quad
Chuang Gan$^{1}$ \quad
\\
\vspace{-0.5em}
\\
$^1$UMass Amherst\quad
$^2$HKUST\quad
$^3$Harvard University\quad\\
}

\begin{document}
\twocolumn[{
    \renewcommand\twocolumn[1][]{#1}
    \maketitle
    \vspace{-10mm}
    \begin{center}{
        \hypersetup{urlcolor=red}
        \url{https://TesserActWorld.github.io}
    }\end{center}
    \centering
    \begin{minipage}{0.95\textwidth}
        \centering
        \includegraphics[trim=000mm 000mm 000mm 000mm, clip=False, width=\linewidth]{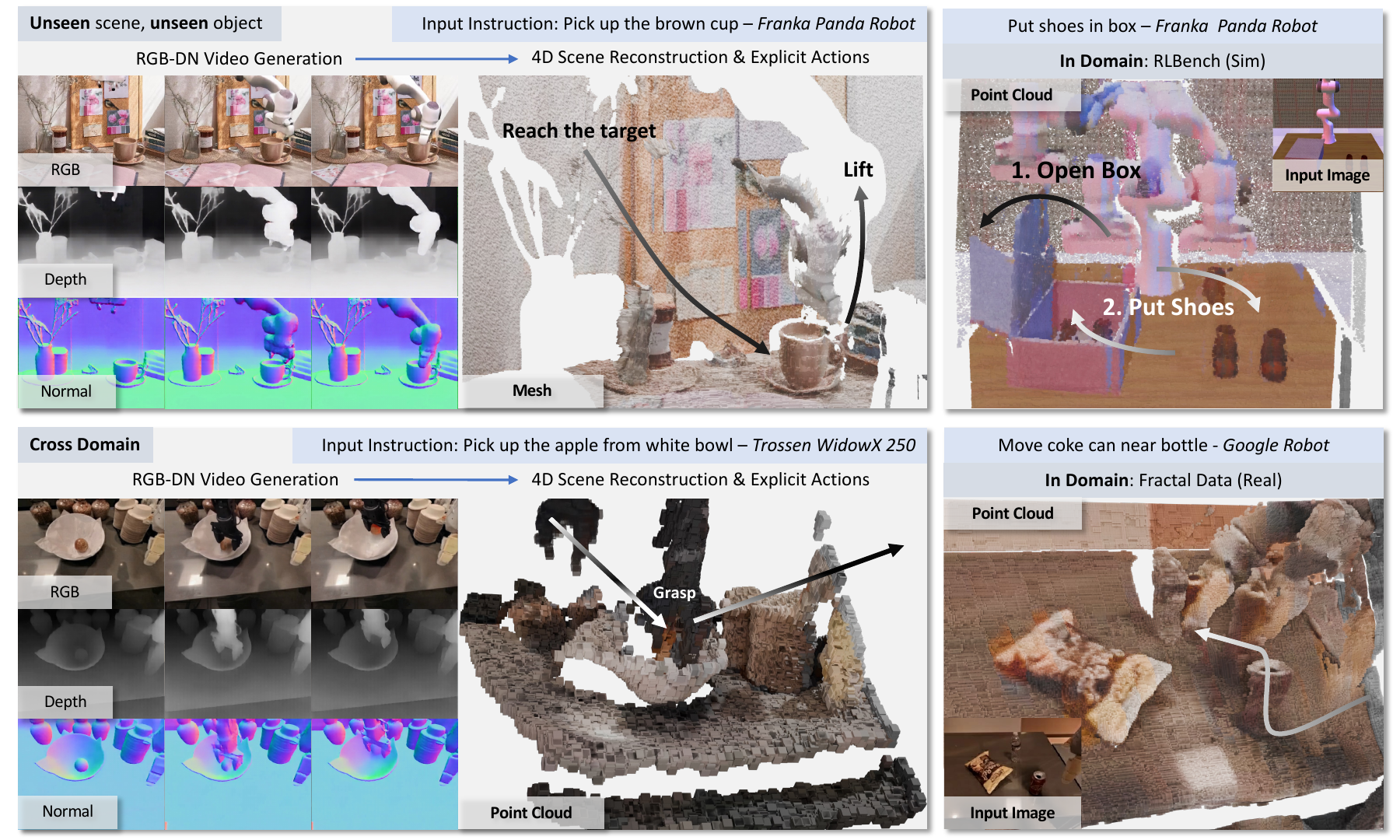}
    \end{minipage}
    \captionsetup{type=figure}
    \vspace{-2mm}
    \captionof{figure}{We propose \modelname, the 4D Embodied World Model, which takes an input image and text instruction to generate RGB, depth, and normal videos, reconstructing a 4D scene and predicting actions. Our model not only achieves strong performance on in-domain data (right) but also generalizes effectively to unseen scenes, novel objects (top left), and cross-domain scenarios (bottom left).}
    \label{fig:teaser}
    \vspace{1em}
}]

\maketitle
\renewcommand{\thefootnote}{}
\footnotetext{$^*$Equal contribution.}

\begin{abstract}
This paper presents an effective approach for learning novel 4D embodied world models, which predict the dynamic evolution of 3D scenes over time in response to an embodied agent's actions, providing both spatial and temporal consistency. 
We propose to learn a 4D world model by training on RGB-DN (RGB, Depth, and Normal) videos.
This not only surpasses traditional 2D models by incorporating detailed shape, configuration, and temporal changes into their predictions, but also allows us to effectively learn accurate inverse dynamic models for an embodied agent. 
Specifically, we first extend existing robotic manipulation video datasets with depth and normal information leveraging off-the-shelf models. 
Next, we fine-tune a video generation model on this annotated dataset, which jointly predicts RGB-DN (RGB, Depth, and Normal) for each frame. 
We then present an algorithm to directly convert generated RGB, Depth, and Normal videos into a high-quality 4D scene of the world. Our method ensures temporal and spatial coherence in 4D scene predictions from embodied scenarios, enables novel view synthesis for embodied environments, and facilitates policy learning that significantly outperforms those derived from prior video-based world models.

\end{abstract}

\vspace{-1em}
\section{Introduction}
Learned world models~\citep{ha2018world, yang2023learning, opensora, xiang2024pandora}, which simulate environmental dynamics, play a crucial role in enabling embodied intelligent agents.
Such models enable flexible policy synthesis \citep{du2024learning, liang2024dreamitate}, data simulation and generation \citep{yang2023learning, zhu2024irasim}, and long-horizon planning \citep{janner2022planning, du2023video, zhang2024combo}. 
However, while the physical world is three-dimensional in nature, existing world models operate in the space of 2D pixels.
This limitation leads to an incomplete representation of spatial relationships, impeding tasks that require precise depth and pose information. For instance, without accurate depth and 6-DoF pose estimations, robotic systems struggle to determine the exact position and orientation of objects. Furthermore, existing 2D models can produce unrealistic results, such as inconsistent object sizes and shapes across time steps, which limits their use in data-driven simulations and robust policy learning.

In this paper, we explore how we can instead learn a 4D embodied world model, \modelname, which simulates the dynamics of a 3D world.
This 4D embodied world model allows us to generate realistic 3D interactions, such as grasping objects or opening drawers, with a level of detail that traditional 2D-based models cannot achieve. By modeling spatial and temporal dimensions, our model provides the depth and pose information essential for robotic manipulation.
However, the task of learning a 4D embodied world model is challenging as the dynamics of the world are extremely computationally expensive to train and learn, requiring models to generate outputs in three-dimensional space and time. To efficiently represent and predict the dynamics of the world, we propose a substantially more lightweight representation of the 4D world, consisting of predicting a sequence of RGB, depth, and normal maps of the scene. This combined representation accurately captures the appearance, geometry, and surface of a scene while being substantially lower dimensional than explicitly predicting world dynamics. Furthermore, such a representation shares substantial similarities to existing video models, allowing us to directly use the generative capabilities of existing video models to effectively construct our 4D world model.

Given this intermediate representation, we present an efficient algorithm to reconstruct accurate 4D scenes from generated RGB-DN videos. 
For each frame, we use a combination of both depth and normal prediction to integrate a smooth 3D surface of the scene. We then use optical flow between generated frames to distinguish between background and dynamic regions in the reconstructed 3D scene across frames and introduce two novel loss functions to enforce consistency across scenes over time. As shown in Figure~\ref{fig:teaser}, our 4D Embodied World Model enables the construction of high-fidelity 4D-generated scenes, facilitating strong-performance action planning for downstream tasks.

A key challenge for training \modelname is a lack of access to existing large-scale datasets with high-quality 4D annotations, or the RGB image, depth, and normal information needed to train our approach. To overcome this, we collect a 4D embodied video dataset consisting of synthetic data with ground truth depth and normal information and real-world data with estimated depth and normal information with off-the-shelf estimators.

Overall, our paper has the following contributions:
\begin{itemize}
    \item We collect a 4D embodied video dataset with compact and high-quality RGB, depth, and normal annotations and learn a 4D embodied world model.
    \item We present an algorithm to convert the generated RGB-DN video into high-quality 4D scenes coherent across both time and space.
    \item Extensive experiments demonstrate our 4D embodied world model can predict high-fidelity 4D scenes and achieve superior performance in downstream embodied tasks compared to traditional video-based world models.
\end{itemize}

\begin{table*}[ht]
\centering
\scalebox{1.0}{
\begin{tabular}{l|cccccc}
\toprule
Dataset & Domain & Depth Source & Normal Source & Embodiment & \# of videos \\ 
\midrule
RLBench~\citep{james2019rlbench} & Synthetic & Simulator & Depth2Normal~\citep{bae2024dsine} & Franka Panda & 80k \\
\midrule
RT1 Fractal Data~\citep{brohan2022rt} & \multirow{3}{*}{Real} & \multirow{3}{*}{Rolling Depth~\citep{ke2024rollingdepth}} & \multirow{3}{*}{Marigold~\citep{ke2024repurposing}} & Google Robot & 80k \\ 
Bridge~\citep{walke2023bridgedata} & & & & WidowX & 25k \\ 
SomethingSomethingV2~\citep{goyal2017something} & & & & Human Hand & 100k \\ 

\bottomrule
\end{tabular}
}
\caption{\textbf{Overview of the 4D embodied video datasets.}}
\label{tab:data}
\end{table*}

\section{Related Work}

\textbf{Embodied Foundation Models}
A flurry of recent work has focused on constructing foundation models for general purpose agents~\citep{yang2023foundation, firoozi2023foundation}. One line of work has focused on constructing multimodal language models that operate over images~\citep{li2022pre, jiang2022vima, raman2022planning, driess2023palm, wang2023describe, zhang2023building,gramopadhye2022generating} as well as 3D inputs~\citep{hong20233d, huang2023embodied} and output text describing the actions of an agent. Other works have focused on the construction of vision-language-action (VLA) models that directly output action tokens~\citep{brohan2023rt, kim2024openvla, zhen20243d}. Both of the previous approaches aim to construct foundation model policies (over text or continuous actions). In contrast, our work aims to instead construct a foundation 4D world model for embodied agents, which can then be used for downstream applications such as planning~\citep{du2023video, zhang2024combo} or policy synthesis~\citep{du2024learning, liang2024dreamitate}.

\noindent\textbf{Learning World Models}
Learning dynamics model of the world given control inputs has been a long-standing challenge in system identification~\citep{ljung1994modeling}, model-based reinforcement learning~\citep{sutton1991dyna}, and optimal control~\citep{AstromWittenmark1973, bertsekas1995dynamic}. A large body of work focused on learning world models in the low dimensional state space~\citep{ferns2004metrics,achille2018separation,lesort2018state}, which while being efficient to learn, is difficult to generalize across many environments. Other works have explored how world models may be learned over pixel-space images~\citep{chiappa2017recurrent, ha2018world, hafner2021mastering, chen2022transdreamer, micheli2022transformers}, but such models are trained on simple game environments. With advances in generative modeling, a large flurry of recent research has focused on using video models as foundation world models~\citep{yang2023learning, openai2023videogeneration, opensora, xiang2024pandora, bruce2024genie, zhou2024robodreamer} but such models operate over the space of 2D pixels which does not fully simulate the 3D world. Most similar to our work are \citep{zhen20243d}, which predicts only the 3D goal state for robotic tasks, and \cite{team2025aether}, an geometric-aware world model trained purely on synthetic data without language grounding or downstream robotic tasks; in contrast, our approach models the 4D scene from RGB-DN videos and supports language-conditioned control.

\noindent\textbf{4D Video Generation} The task of 4D video generation has gained increasing attention in recent years~\citep{4dgen_yin20234dgen,t24d_singer2023text}, driven by advancements in diffusion models \citep{ddpm_ho2020denoising, rombach2022high, peebles2023scalable}, neural radiance fields \citep{nerf_mildenhall2021nerf}, and 3D Gaussian splatting \citep{3dgs_kerbl20233d}. However, existing methods often suffer from slow optimization due to hybrid frameworks \citep{4dfy_bahmani20244d, sv4d_xie2024sv4d, animate124_zhao2023animate124, ayg_ling2024align} and the convergence challenges of SDS loss \citep{consistent4d_jiang2023consistent4d, dreamgaussian4d_ren2023dreamgaussian4d}. Instead, we represent 4D scenes using RGB-DN videos, which offer more efficient training and provide high-accuracy 3D information crucial for embodied tasks. Furthermore, our approach is the first to directly predict 4D scenes from the current frame and the embodied agent’s action described in the text.

\section{Preliminaries}

\subsection{Latent Video Diffusion Models}

Diffusion models \citep{ddpm_ho2020denoising, rombach2022high} are capable of learning the data distribution $p(x)$ by progressively adding noise to the data until it resembles a Gaussian distribution through a forward process. During inference, a denoiser $\epsilon$ is trained to recover the data from this noisy state. Latent video diffusion models \citep{opensora} utilize a Variational Autoencoder (VAE) \cite{kingma2013auto, vqvae_van2017neural} to encode the data in the latent space, maintaining high-quality outputs while more efficiently modeling the data distribution. 

We formulate the task of RGB $\mathcal{V}$, depth $\mathcal{D}$, and normal $\mathcal{N}$ video generation as a conditional denoising generation task, i.e., we model the distribution $p(\mathbf{v}, \mathbf{d}, \mathbf{n} | \mathbf{v}^0, \mathbf{d}^0, \mathbf{n}^0, \mathcal{T})$, where $\mathbf{v}, \mathbf{d}, \mathbf{n}$ represent the predicted future latent sequences of RGB, depth, and normal maps, respectively; the conditions $\mathbf{v}^0, \mathbf{d}^0, \mathbf{n}^0,\mathcal{T}$ are the latent of RGB image, depth and normal maps, and embodied agent's action in text.

The forward diffusion process adds Gaussian noise to the latent $\mathbf{z}\in\{\mathbf{v}, \mathbf{d}, \mathbf{n}\}$ over $T$ timesteps, defined as: 
\begin{equation}
    q(\mathbf{z}_t | \mathbf{z}_{t-1}) = \mathcal{N}\left(\mathbf{z}_t; \sqrt{\alpha_t} \mathbf{z}_{t-1}, (1 - \alpha_t) \mathbf{I}\right)
    \label{eq:noise}
\end{equation}
where $t \in \{1, 2, \dots, T\}$ denotes the diffusion step, $\alpha_t$ is a parameter controlling the noise influence at each step, and $\mathbf{I}$ is the identity matrix. In the reverse process, the model aims to recover the original latent from the noise.  Let $\mathbf{x}=[\mathbf{v}, \mathbf{n}, \mathbf{d}]$ denoting the concatenation of $\mathbf{v}, \mathbf{n}, \mathbf{d}$, a denoising network $\epsilon_\theta(\mathbf{x}_t, t, \mathbf{x}^0, \mathcal{T})$ with learning parameters $\theta$ is trained to predict the noise added at each timestep. The reverse process is defined as:
\begin{gather}
\resizebox{0.91\linewidth}{!}{$
    p_\theta(\mathbf{x}_{t-1} |\mathbf{x}_t, \mathbf{x}^0, \mathcal{T}) = \mathcal{N}\left(\mathbf{x}_{t-1}; \mu_\theta(\mathbf{x}_t, t, \mathbf{x}^0, \mathcal{T}), \Sigma_\theta(\mathbf{x}_t, t)\right)
    $}
\end{gather}
Once the denoised latent $\mathbf{x}_0$ is obtained, the model maps it back to the pixel space to obtain the final RGB-DN video.

\subsection{Depth Optimization via Normal Integration}
\label{sec:depop}
As discussed in ~\citep{cao2022bilateral, xiu2023econ}, normal maps provide essential information about surface orientation, which is vital for enforcing geometric constraints and imposing surface smoothness and continuity during depth optimization. This spatial optimization leads to more accurate and reliable depth estimates that closely align with the true 3D geometry and capture fine surface details.

To formalize the process, we use the perspective camera model to set constraints on the depth and surface normal. In the coordinate system of the 2D image at frame $i$, a pixel position is given as $\boldsymbol{u}=(u,v)^T\in\mathcal{V}^i$, and its corresponding depth scalar, normal vector is $d\in \mathcal{D}^i, \boldsymbol{n}=(n_x, n_y, n_z)\in \mathcal{N}^i$. Under the assumption of a perspective camera with focal length $f$ and the principal point $(c_u,c_v)^T$, as proposed by~\citep{durou2009integrating}, the log-depth $\tilde{d}=\log(d)$ should satisfy the following equations: $\tilde{n}_z\partial_u\tilde{d}+n_x=0$ and $\tilde{n}_z\partial_v\tilde{d}+n_y=0$ where $\tilde{n}_z=n_x(u-c_u)+n_y(v-c_v)+n_z f$. In addition, we can add the assumption that all locations are smooth surfaces~\citep{cao2022bilateral}. We can convert the above constraint to the quadratic loss function, allowing us to find the optimized depth map:
\begin{equation}
    \min_d \iint_\Omega (\tilde{n}_z\partial_u\tilde{d}+n_x)^2+(\tilde{n}_z\partial_u\tilde{d}+n_y)^2\dd u \dd v.
\end{equation}
Following~\citep{cao2022bilateral}, we can convert the above objective to an iteratively optimized loss objective. At iteration step $t$, we can compute the matrix $W(\tilde{d}_t)$ and iteratively optimize for a refined depth prediction $\tilde{d}_{t+1}$:
\begin{equation*}
\scalebox{0.8}{$
    \tilde{d}_{t+1} =
    \argmin_{\tilde{d}}
    (A\tilde{d}-b)^TW(\tilde{d}_t)(A\tilde{d}-b)
    \stackrel{\text{def}}{=}
    \argmin_{\tilde{\mathcal{D}}} \mathcal{L}_{s}(\tilde{\mathcal{D}}, \mathcal{N}^i)$
}
    \label{eq:bini-update}
\end{equation*}
where $A$ and $b$ are defined by normals and camera intrinsic.

\begin{figure*}[t]
    \centering
    \includegraphics[width=0.98\linewidth]{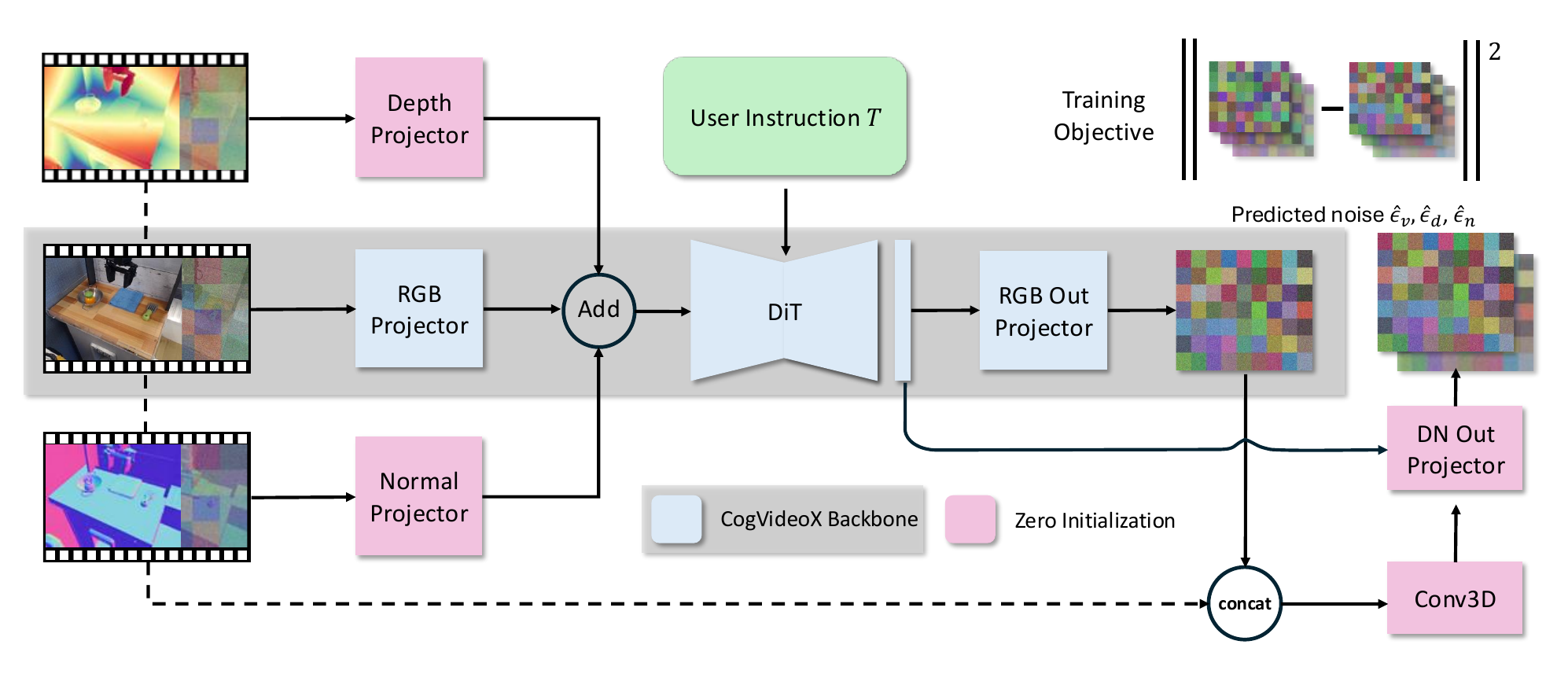}
     \vspace{-4pt}
    \caption{\textbf{Architecture and Training Overview of \modelname.}}
    \label{fig:arch}
\end{figure*}

\section{Learning a 4D Embodied World Model}
Learning how 3D scenes may change over time based on the current observation and action is crucial for embodied agents. We propose to learn a 4D embodied world model, \modelname, by training on RGB-DN videos and reconstructing the 4D scenes from it. We first introduce the 4D embodied video dataset we collected in Sec.~\ref{sec:4d-data}, then discuss the model architecture and training strategy in Sec.~\ref{sec:rgbdn}.
In Sec.~\ref{sec:4d-scene}, we propose an efficient optimization algorithm with two novel loss functions to convert the generated RGB-DN videos into 4D scenes. Finally, in Sec.~\ref{sec:inv-dyn}, we demonstrate how the 4D world model can help downstream embodied tasks. 

\subsection{4D Embodied Video Dataset}
\label{sec:4d-data}

Learning 4D embodied world models requires large-scale 4D datasets, which are expensive to collect in the real world. In this section, we present a data collection and annotation pipeline that enables us to automatically construct 4D datasets from existing video datasets.

Simulator-synthesized data provide ground truth depth information, so we first selected 20 tasks of relatively high difficulty from RLBench~\citep{james2019rlbench} and generated 1000 instances captured from 4 different views for each task, making 80k synthetic 4D embodied videos in total. Although the simulator provides metric depth information, it lacks surface normal data. To estimate normals, we use the \texttt{depth2normal} function from DSINE~\cite{bae2024dsine}. To enhance the generalization, we adopt the scene randomization techniques from the Colosseum data generation pipeline~\cite{pumacay2024colosseum}, which alters the background, table texture and light of the scene.

While synthetic data provides depth and normal data of high quality, their diversity is limited, resulting in a gap compared to real-world scenarios. To bridge this gap, we also incorporate real-world video datasets. As most of these datasets lack depth and normal annotations, we employ the state-of-the-art video depth estimator RollingDepth~\citep{ke2024rollingdepth} to annotate the videos with affine-invariant depth. As affine-invariant depth map does not directly yield normal map as metric depth does and a reliable video normal estimator is currently unavailable, we annotate normal maps using Temporal-Consistent Marigold-LCM-normal$^1$\footnote{\url{https://huggingface.co/docs/diffusers/en/using-diffusers/marigold_usage\#frame-by-frame-video-processing-with-temporal-consistency}}. These two approaches allow us to obtain high-quality, sharp, and temporally consistent video depth and normal annotations.
Specifically, we select two high-quality datasets from OpenX~\citep{open_x_embodiment_rt_x_2023}, the Fractal data~\citep{brohan2022rt}, and the Bridge~\cite{walke2023bridgedata} dataset. Moreover, to further increase the diversity of the instructions, we incorporated the human-object interaction dataset, Something Something V2~\cite{goyal2017something}. Detailed statistics are shown in Table~\ref{tab:data}.

\begin{figure*}[tbp]
    \centering
    \includegraphics[width=0.98\linewidth]{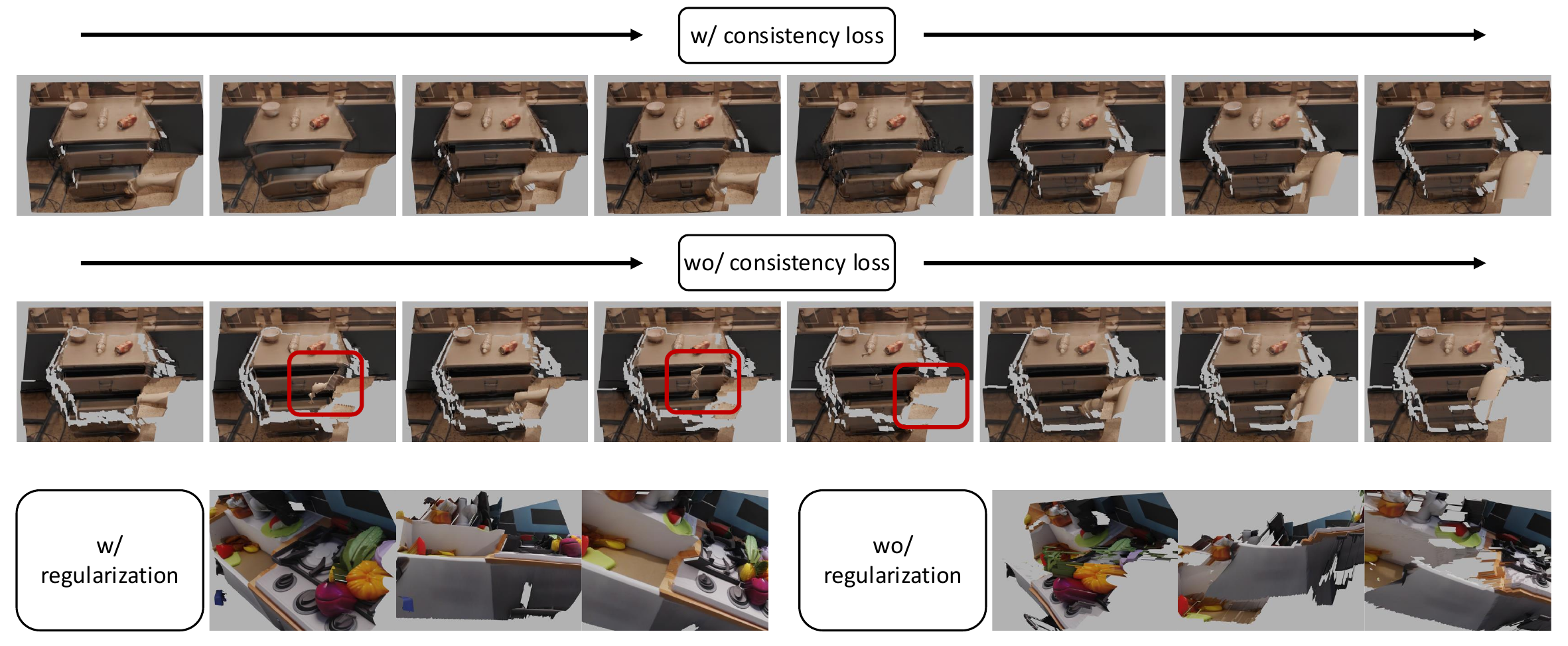}
    \caption{\textbf{Effect of consistency and regularization loss on 4D scene reconstruction.} The red boxes highlight the inconsistent regions.}
    \label{fig:bini-loss}
\end{figure*}

\subsection{Model Architecture and Training Strategy}
\label{sec:rgbdn}

Training a diffusion model to generate temporal RGB-DN data is challenging. To effectively train RGB video models, large-scale video datasets containing billions of high-quality samples are typically employed \citep{opensora, yang2024cogvideox}. In contrast, our RGB-DN dataset, even with automatic annotation, comprises only about 200k data points, which is insufficient to train a world model from scratch. To overcome this limitation, we modify and fine-tune CogVideoX~\citep{yang2024cogvideox} to serve as our RGB-DN prediction model, leveraging the pre-trained knowledge within it to effectively bootstrap our 4D model.

Our architecture is illustrated in Figure~\ref{fig:arch}. First, we utilize the 3D VAE~\cite{kingma2013auto, vqvae_van2017neural} from CogVideoX to separately encode the RGB ($\mathbf{v}$), depth ($\mathbf{d}$), and normal ($\mathbf{n}$) videos, without any additional fine-tuning of the VAE.
These latent representations are perturbed with noise to get $\mathbf{x}_t$, and are then fed into our model along with the corresponding image latent head $\mathbf{x}^0$. For the input design, we introduce three separate projectors for each modality to extract the embeddings: \(f_\mathbf{z}=\texttt{InputProj}(\mathbf{z}_t,\mathbf{z}^0)\), where $\mathbf{z} \in \{\mathbf{v}, \mathbf{d}, \mathbf{n}\}$.
DiT then takes the sum of these embeddings as input, conditioned on the textual input $\mathcal{T}$ and denoising step $t$, to obtain the hidden state: \(\mathbf{h} = \texttt{DiT}\qty(\sum f_\mathbf{z}, t, \mathcal{T})\). To distinguish between different robot arms, $\mathcal{T}$ is defined as [action instruction] + [robot arm name], e.g., ``pick up apple \emph{google robot}''. On the output side, we retain the original RGB output method: $\epsilon^*_\mathbf{v} = \texttt{OutputProj}(h)$. 
However, we introduce additional modules for depth and normal prediction. A Conv3D layer is used to encode the concatenation of the input latents and the predicted RGB denoised output.
These are then combined with the hidden states produced by the DiT backbone and passed through the output projector to obtain the denoised predictions for depth and normal:
\(\epsilon^*_{\mathbf{d},\mathbf{n}} = \texttt{DNProj}(h, \texttt{Conv3D}(\epsilon^*_\mathbf{v}, [\mathbf{z}_t;\mathbf{z}^0]_{z \in \{v,d,n\}}))\).
To preserve the pretrained knowledge, we initialize our model with the CogVideoX weights. All other modules are initialized with zeros, ensuring that the RGB output at the beginning of training matches the output of CogVideoX. During training, we randomly select samples from the 4D embodied dataset $(\mathcal{V}, \mathcal{D}, \mathcal{N}, \mathcal{T})$ constructed above and apply Eq.\ref{eq:noise} to add noise $\epsilon_\mathbf{v}$, $\epsilon_\mathbf{d}$, and $\epsilon_\mathbf{n}$ to the RGB-DN data at timestep $t$, minimizing the following objective:
\begin{equation}
    L =
    \mathbb{E}_{\mathbf{v}_0, \mathcal{T}, t, \epsilon}
    \left[ \bigg\| [\epsilon_\mathbf{v}, \epsilon_\mathbf{d}, \epsilon_\mathbf{n}] - \epsilon_\theta(\mathbf{x}_t, t, \mathbf{x}^0, \mathcal{T}) \bigg\|^2 \right]
\end{equation}

\begin{table*}[t]
\centering
\resizebox{\textwidth}{!}{
\begin{tabular}{ll|ccc|ccc|ccc|c}
\toprule
\multirow{2}{*}{Domain} & \multirow{2}{*}{Method} & \multicolumn{3}{c|}{RGB} & \multicolumn{3}{c|}{Depth} & \multicolumn{3}{c|}{Normal} &  \multicolumn{1}{c}{Point Cloud} \\
\multicolumn{1}{c}{} &  & \multicolumn{1}{c}{FVD $\downarrow$} & \multicolumn{1}{c}{SSIM $\uparrow$} & \multicolumn{1}{c|}{PSNR $\uparrow$} & \multicolumn{1}{c}{AbsRel $\downarrow$} & \multicolumn{1}{c}{$\delta_1$ $\uparrow$} & \multicolumn{1}{c|}{$\delta_2$ $\uparrow$} & \multicolumn{1}{c}{Mean $\downarrow$} & \multicolumn{1}{c}{Median $\downarrow$} & \multicolumn{1}{c|}{$11.25^\circ$ $\uparrow$} & \multicolumn{1}{c}{Chamfer $L_1$ $\downarrow$} \\ \midrule
\multirow{4}{*}{Real} & 4D Point-E & - & - & - & - & - & - & - & - & - & 0.2211 \\
& OpenSora & \multicolumn{1}{c}{23.67} & \multicolumn{1}{c}{71.31} & \multicolumn{1}{c|}{19.25} & \multicolumn{1}{c}{31.41} & \multicolumn{1}{c}{60.39} & \multicolumn{1}{c|}{79.97} & \multicolumn{1}{c}{41.82} & 32.15 & 13.58 & 0.3013 \\
& CogVideoX & \textbf{20.64} & \textbf{79.38} & \textbf{22.39} & \underline{26.17} & \underline{64.82} & \underline{81.62} & \underline{19.53} & \underline{10.09} & \underline{22.70} & \underline{0.2191} \\
& \modelname (Ours) & \underline{21.59} & \underline{75.86} & \underline{20.27} & \textbf{22.07} & \textbf{66.80} & \textbf{82.60} & \textbf{15.74} & \textbf{7.32} & \textbf{27.80} & \textbf{0.2030}\\
\midrule
\midrule
\multirow{4}{*}{Synthetic} & 4D Point-E & - & - & - & - & - & - & - & - & - & \underline{0.1086} \\
& OpenSora &  \multicolumn{1}{c}{54.11} & \multicolumn{1}{c}{65.90} & \multicolumn{1}{c|}{19.28} & \multicolumn{1}{c}{\underline{18.40}} & \multicolumn{1}{c}{\underline{65.02}} & \multicolumn{1}{c|}{\underline{91.20}} & \multicolumn{1}{c}{\textbf{12.94}} & \underline{7.58} & 25.02 & 0.2570\\
& CogVideoX & \underline{41.23} & \underline{76.60} & \textbf{20.87} & 19.81 & 60.07 & 80.16 & 20.36 & 10.47 & \underline{26.04} & 0.2884 \\
& \modelname (Ours) & \textbf{40.01} & \textbf{77.59} & \underline{19.73} & \textbf{16.02} & \textbf{69.26} & \textbf{93.03} & \underline{14.75} & \textbf{6.34} & \textbf{36.85} & \textbf{0.0811} \\
\bottomrule
\end{tabular}
}
\caption{\textbf{Main results on the 4D scene generation.} All metrics are averaged over 10 runs for each of the samples. The best results are in \textbf{bold}, and the second best are in \underline{underlined}. \modelname  predicts the depth and normal maps most accurately without hurting RGB much and thus achieves the best accuracy of reconstructed 4D point clouds across real and synthetic image domains.}
\label{tab:4dgen}
\end{table*}

\subsection{4D Scene Reconstruction}
\label{sec:4d-scene}
After obtaining the RGB-DN video, we further optimize the depth and reconstruct the 4D scene. Similar to prior works~\citep{ye2024stablenormal, ke2024repurposing}, our depth representation for each image is given by a relative map in the range $[0, 1]$, and thus cannot directly reconstruct the entire scene. While past work has sidestepped this by assuming either a default scale for depth~\citep{yu2024wonderworld} or by directly predicting metric depth~\citep{zhen20243d, Chen_2024_CVPR}, such reconstructions from depth are often coarse and often cause reconstructed planes or walls to be tilted. 

With the normal maps $\mathcal{N}^i$, we can optimize the depth maps $\mathcal{D}^i$ via normal integration for refined depth maps $\hat{\mathcal{D}}^i$ as introduced in Sec.~\ref{sec:depop} with a loss term $\mathcal{L}_s$ for spatial consistency. However, this approach optimizes depth frame by frame, which lacks temporal consistency across the dynamic scene. To address this, we compute optical flow between frames ~\citep{teed2020raft} $\mathcal{F} = \text{RAFT}(\mathcal{V})$ and enforce consistency of depth across frames.
We define the static regions of each frame as the pixels with the magnitude of optical flow smaller than threshold $c$ and obtain its mask by $\mathcal{M}_s^i = \|\mathcal{F}^i\| \leq c$. We then define the dynamic parts of an image as $\mathcal{M}^i_d=\neg {\mathcal{M}}^i_s$. We further define the background of an image as static regions that are fixed across image frames, $\mathcal{M}^i_b = \mathcal{M}_s^i \cap \mathcal{M}_s^{i-1}$

Since optical flow represents the movement of objects in the 2D-pixel space, we can retrieve the depth at any position from the previous frame to impose consistency constraints. To compute the depth values from the previous frame at positions corresponding to the current frame, we utilize the optical flow $\mathcal{F}^{i \rightarrow (i-1)}$. For each pixel $(u, v)$ in frame $i$, the optical flow provides the displacement $(\Delta u, \Delta v)$, allowing us to find the corresponding pixel in frame $i-1$ at position $(u - \Delta u, v - \Delta v)$. Based on this mapping, we define the $\mathcal{D}^{i\rightarrow (i-1)}$ such that: $\mathcal{D}^{i\rightarrow (i-1)}(u, v) = \mathcal{D}^{i-1}(u - \Delta u, v - \Delta v)$.
We then introduce the consistency loss over the dynamic and background region of the image $\mathcal{L}_c$:
\begin{equation}
\scalebox{0.8}{
    $\begin{aligned}
        \mathcal{L}_c(\tilde{\mathcal{D}}, \hat{\mathcal{D}}^{i-1}, \mathcal{F}^i, \mathcal{F}^{i-1}) =&
        \lambda_{cd} \left\| \tilde{\mathcal{D}}^i \circ \mathcal{M}^i_d - \mathcal{D}^{i\rightarrow (i-1)} \circ \mathcal{M}^i_d \right\|^2 +
        \\ &
        \lambda_{cb} \left\| \tilde{\mathcal{D}}^i \circ \mathcal{M}^i_b - \mathcal{D}^{i\rightarrow (i-1)} \circ \mathcal{M}^i_b \right\|^2
    \end{aligned}$
}
\end{equation}
In addition to the consistency loss, we also incorporate the regularization loss $\mathcal{L}_r$, enforcing that optimized depths are similar to the generated depth map $\mathcal{D}^i$:
\vspace{-3mm}
\begin{equation}
\scalebox{0.67}{
    $\begin{aligned}
    \mathcal{L}_r(\tilde{\mathcal{D}}, \mathcal{D}^{i}) &=
    \lambda_{rd} \left\| \tilde{\mathcal{D}}^i \circ \mathcal{M}^i_d - \mathcal{D}^{i} \circ \mathcal{M}^i_d \right\|^2 +
    \lambda_{rb} \left\| \tilde{\mathcal{D}}^i \circ \mathcal{M}^i_b - \mathcal{D}^{i} \circ \mathcal{M}^i_b \right\|^2
    \end{aligned}$
}
\end{equation}

The overall loss objective we optimize is given by:
\begin{equation}
\label{eq:our-bini}
    \scalebox{0.8}{
    $\argmin_{\tilde{\mathcal{D}}}\ \ 
    \mathcal{L}_s(\tilde{\mathcal{D}}, \mathcal{N}^i) + 
    \mathcal{L}_c(\tilde{\mathcal{D}}, \hat{\mathcal{D}}^{i-1}, \mathcal{F}^i, \mathcal{F}^{i-1}) + 
    \mathcal{L}_r(\tilde{\mathcal{D}}, \mathcal{D}^{i})$
}
\end{equation}

Starting from the first frame, we iteratively refine the depth map by optimizing the loss above, and initialize the depth map at frame $i$ with the generated depth map $\tilde{D}_0 = \mathcal{D}^i$. With refined depth maps $\tilde{\mathcal{D}}$ and RGB Images $\mathcal{V}$, we can reconstruct 4D point clouds representing the world that are consistent over both space and time.

\subsection{Embodied Action Planning with 4D Scenes}
\label{sec:inv-dyn}
After generating 4D scenes, which encapsulate both spatial and temporal information, we extract geometric details that can significantly enhance downstream tasks in robotics. The detailed geometry captured by these scenes plays a crucial role in tasks including robotic grasping.

To achieve this, we employ an inverse dynamics model built on the 4D point clouds, predicting the appropriate robot action $a_i$ based on the current state $s_i$, the predicted future state $s_{i+1}$ and the instruction $\mathcal{T}$. Mathematically, this relationship is expressed as $a_i = \text{ID}(s_i, s_{i+1}, \mathcal{T})$, where 
$s_i$ represents the scene at the time step $i$. Specifically, we use PointNet~\citep{qi2017pointnet++} to encode the point cloud and extract 3D features. These features are then combined with the instruction text embeddings and further processed by an MLP to generate the final 7-DoF action.

\begin{table*}[ht]
\centering
\small
\resizebox{\linewidth}{!}{
\begin{tabular}{lcccccccccccc}
\toprule
Methods & 
\makecell{\small\texttt{close}\\\small\texttt{box}} & 
\makecell{\small\texttt{open}\\\small\texttt{drawer}} & 
\makecell{\small\texttt{open}\\\small\texttt{jar}} & 
\makecell{\small\texttt{open}\\\small\texttt{microwave}} & 
\makecell{\small\texttt{put}\\\small\texttt{knife}} & 
\makecell{\small\texttt{sweep to}\\\small\texttt{dustpan}} & 
\makecell{\small\texttt{lid}\\\small\texttt{off}} & 
\makecell{\small\texttt{weighing}\\\small\texttt{off}} & 
\makecell{\small\texttt{water}\\\small\texttt{plants}} \\
\midrule
Image-BC & 53 & 4 & 0 & 5 & 0 & 0 & 12 & 21 & 0 \\
UniPi$^*$   & 81 & 67 & 38 & \textbf{72} & 66 & 49 & 70 & \textbf{68} & 35 \\
4DWM (Ours) & \textbf{88} & \textbf{80} & \textbf{44} & 70 & \textbf{70} & \textbf{56} & \textbf{73} & 62 & \textbf{41} \\
\bottomrule
\end{tabular}
}
\caption{\textbf{\modelname boosts the performance of action planning.} We report the success rate averaged over 100 episodes for each task here.}
\label{tab:exp-rlbench}
\end{table*}

\begin{table}[t]
\centering
\resizebox{\linewidth}{!}{
\begin{tabular}{lcccccc}
\toprule
Method      & PSNR      & SSIM      & LPIPS         & \makecell{CLIP\\Score}        & \makecell{CLIP\\Aesthetic} & \makecell{Time\\Costs}\\
\midrule
SoM\cite{wang2024shape} &   10.94    &    24.02       &        \textbf{73.82}       &  66.67  &     3.61          & $\sim$2 hours \\
Ours       &   \textbf{12.99}    &     \textbf{42.62}      &     60.51          &  \textbf{83.02}  & \textbf{3.73} & $\sim$ \textbf{1 mins} \\
\bottomrule
\end{tabular}
}
\caption{\textbf{Novel view synthesis results on RLBench.}}
\label{tab:nvs}
\end{table}

\section{Experiments}
We first evaluate the quality of 4D scene predictions from our model across real and synthetic datasets in Sec.~\ref{sec:exp-4d}, then conduct experiments in RLBench to demonstrate how the 4D information helps the embodied tasks in Sec.~\ref{sec:exp-action}. We provide more qualitative results and videos in the Supplementary and the website.

\subsection{4D Scene Prediction}
\label{sec:exp-4d}

\subsubsection{Setup}
\noindent\textbf{Datasets.} We conduct experiments on the real domain with 400 unseen samples in RT1 Fractal~\citep{brohan2022rt} and Bridage~\citep{walke2023bridgedata} dataset where depth and normal are estimated as in Sec.~\ref{sec:4d-data}, and the synthetic domain with 200 unseen samples in RLBench~\citep{james2019rlbench}, where ground truth depth and normal maps are directly accessible.

\noindent\textbf{Metrics.} We evaluate the video quality with FVD, SSIM, and PSNR; depth quality with AbsRel, $\delta_1$, and $\delta_2$; normal maps quality with Mean, Median, and $11.25^\circ$; reconstructed point cloud quality with the $L_1$ Chamfer Distance.
We generate 10 times per sample and report the average.

\noindent\textbf{Baselines.} We compare our method to two video diffusion models and a 4D point cloud diffusion model. %
\begin{itemize}
[align=right,itemindent=0em,labelsep=3pt,labelwidth=0em,leftmargin=1em,itemsep=0em]
\item OpenSora~\citep{opensora}, video diffusion model fine-tuned with LoRA on the same dataset without depth and normal annotations. To construct the full 4D scene, depth and normal are additional estimated given the predicted video with Rolling Depth and Marigold.
\item CogVideoX~\citep{yang2024cogvideox}, video diffusion model fine-tuned with LoRA on the same dataset without depth and normal annotations. The 4D scene is obtained similarly to the above.
\item 4D Point-E, since no prior work directly generates dynamic scenes from the first frame and text inputs, we implemented a 4D point cloud diffusion model as the baseline, where we modify the Point-E \citep{nichol2022point} model by conditioning it on the mean of CLIP \citep{radford2021learning} features extracted from both text and image inputs, outputting $T$ point clouds of size $n$, where T is set to 4 and $n$ is set to $8192$ due to computational constraints.
\end{itemize}

\noindent\textbf{Implementation Details.}
We train our model on the collected 4D embodied video dataset using a multi-resolution training approach, predicting 49 frames at a time. For more details, please kindly refer to the Supplementary Materials.
\begin{figure*}[tbp]
    \centering
    \includegraphics[width=0.99\linewidth]{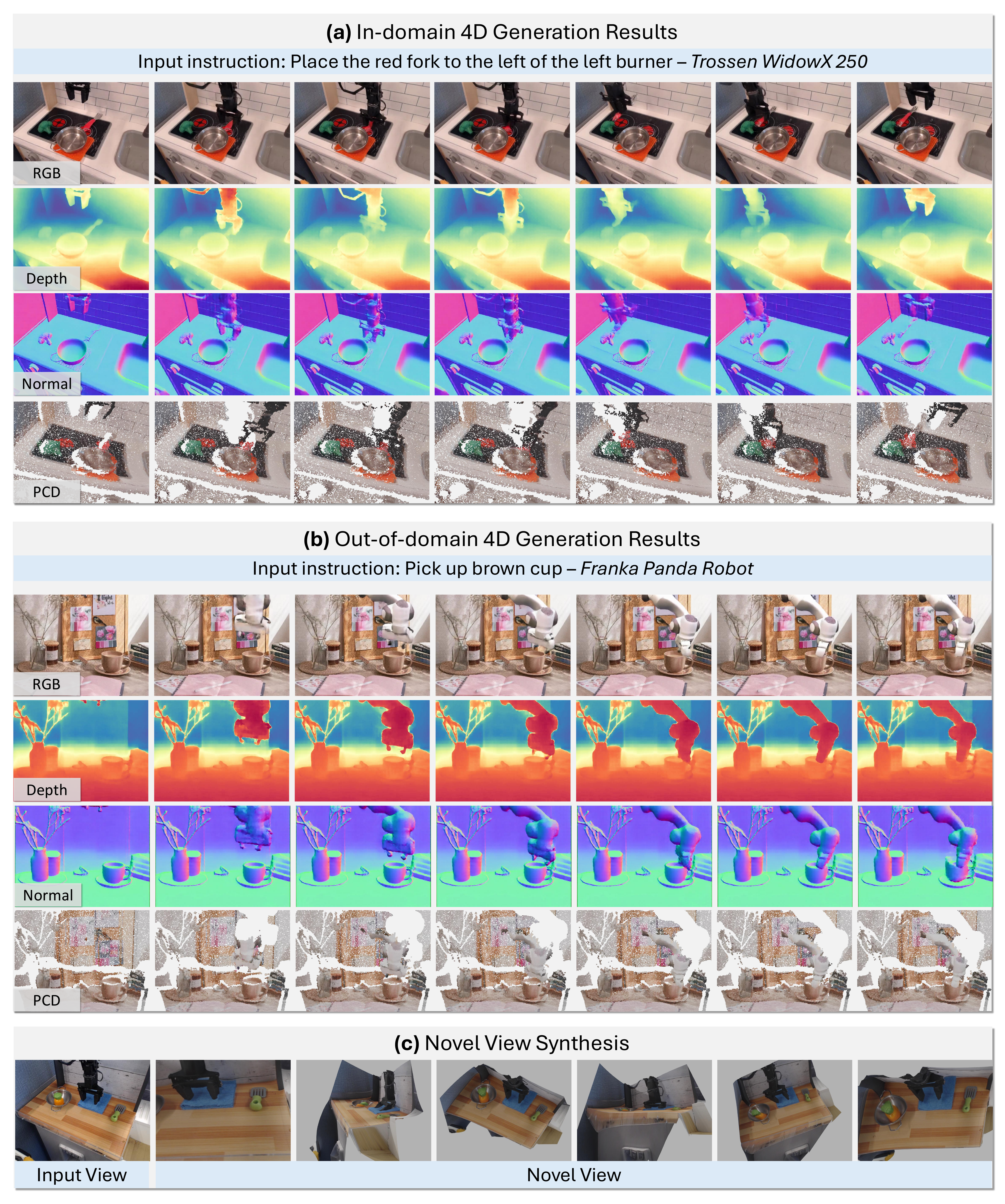}
    \caption{\textbf{Qualitative results of} (a) \textbf{In-domain 4D generation results.} (b) \textbf{Generalization over unseen scenes and objects.} (c) \textbf{Novel view synthesis.}}
    \label{fig:examples}
\end{figure*}
\subsubsection{Results and Analysis}

\noindent\textbf{\modelname predicts high quality 4D scenes.}
As shown in Table~\ref{tab:4dgen}, our model accurately predicts the depth and normal maps compared to video diffusion models with post-estimation, verifying the effectiveness of our learned model. With better depth and normal maps, the 4D point clouds reconstructed with our method achieve the lowest Chamfer distances across real and synthetic datasets. 
The 4D Point-E method performs better than video diffusion models, particularly on RLBench, but still lags behind our approach. Additionally, training directly with point clouds is computationally expensive, restricting the number of frames used. In contrast, our model leverages an efficient representation with RGB-DN videos to generate more precise 4D scenes, particularly in capturing fine-grained details in dynamic scenes. We also show qualitative results in Figure~\ref{fig:examples} (a).

\noindent\textbf{\modelname synthesizes novel views efficiently.}
\label{sec:exp-nvs}
Our method could also perform monocular video to 4D tasks by predicting depth and normal sequences and generating point clouds. We conduct experiments on the task of novel view synthesis on RLBench and compare with Shape of Motion \citep{wang2024shape}, a state-of-the-art video reconstruction approach that utilizes Gaussian splatting \citep{3dgs_kerbl20233d}. The input is a monocular front camera video, and we compare results from the overhead and left shoulder cameras. We report PSNR (reconstruction accuracy), SSIM (structural similarity), LPIPS (perceptual difference), CLIP Score (semantic match)~\citep{taited2023CLIPScore}, CLIP aesthetic (visual quality)~\citep{aesthetic-predictor}, and Time costs in Table~\ref{tab:nvs}. The results show our method can synthesize novel views of higher visual quality and better alignment in significantly less time. A qualitative result on the Bridge dataset is shown in Figure~\ref{fig:examples} (c).

\noindent\textbf{Consistency and Regularization Loss are effective.} We conduct ablation experiments on our newly designed loss terms in Sec.~\ref{sec:4d-scene}. The results are shown in Figure~\ref{fig:bini-loss}. The first two rows demonstrate the effect of the consistency loss, where we render frames from the same camera view at different time steps. The results show that the robot arm's movements are more coherent with the consistency loss applied. The last row highlights the role of the regularization loss. We display images of the same frame from three different views, revealing that this loss term helps improve the geometric accuracy of the reconstruction.

\noindent\textbf{\modelname shows generalization across scenes and embodiments.} Our model demonstrates strong generalization capabilities. Benefiting from the knowledge of CogVideoX, the model achieves good generation on unseen scenes and unseen objects. Additionally, it performs well in cross-embodiment scenarios, such as using the Bridge dataset robotic arm in the RT-1 environment. Figure~\ref{fig:examples} (b) shows a generalization result on unseen scenes and objects. More results can be found in the Supplementary Materials.

\subsection{Embodied Action Planning}
\label{sec:exp-action}
\noindent\textbf{Dataset.} We select 9 challenging tasks from RLBench \citep{james2019rlbench} including tasks requiring high-precision grasping.

\noindent\textbf{Metric.} The success rate averaged over 100 episodes.

\noindent\textbf{Baseline.} We compare our method to a behavior cloning agent and a video-based world model.
\begin{itemize}
[align=right,itemindent=0em,labelsep=3pt,labelwidth=0em,leftmargin=1em,itemsep=0em]
    \item Image-BC~\citep{jang2022bc}: a behavior cloning agent that takes in an image and task instruction and outputs the 7-DoF actions.
    \item UniPi$^*$~\citep{du2024learning}: a method that takes the task instruction and current image, predicts the future video, and uses a 2D-based inverse dynamic policy to predict actions. For this baseline, we re-implement it by replacing the backbone with fine-tuned CogVideoX~\citep{yang2024cogvideox} for fair comparison. 
\end{itemize}

\noindent\textbf{Implementation Details.}
We collected 500 samples for each task to train the inverse dynamic model.
Given an initial state during inference, we first predict and record all future keyframes. In subsequent actions, we only query the inverse dynamic model to obtain the corresponding actions by the current state and the predicted future state. We post-trained the model from Sec.~\ref{sec:exp-4d}, allowing the model to predict only the keyframes for each task in RLBench. Our maximum frame length is 13, with a fixed resolution of 512 $\times$ 512.

\subsubsection{Results and Analysis}

The results are shown in Table~\ref{tab:exp-rlbench}, where our method outperforms video diffusion models and image behavior cloning agents in most of the tasks. This is because, in most tasks, 4D  point clouds can reveal the geometry of objects, providing better spatial guidance for robotics planning, as seen in tasks like \texttt{close box} and \texttt{open jar}. At the same time, 3D information can assist with tool use, such as in tasks like \texttt{sweep to dustpan} and \texttt{water plants}. However, in the \texttt{open microwave} and \texttt{weighing off} tasks, the performance is not as good as the baseline, possibly because these tasks already have sufficient information in the 2D front image. Overall, these results highlight the potential of combining 4D scene prediction with inverse dynamic models to improve robotics task execution.

\section{Conclusion}
We learn a 4D generative world model, TesserAct, using a collected 4D embodied video dataset, which consists of robotic manipulation videos annotated with depth and normal information. To ensure both temporal and spatial consistency in scene reconstruction, we introduce two novel loss terms. Our experiments across synthetic and real-world datasets demonstrate that our model generates high-quality 4D scenes and significantly enhances the performance of downstream embodied tasks by leveraging 3D information. We believe that such world models will become increasingly powerful and essential, serving as a foundation for simulating the physical world and advancing the development of intelligent embodied agents. These models will enable fully offline policy training in the real world and facilitate planning through imagined rollouts within the learned world representation.

\section{Limitations}
While our RGB-DN representation of a 4D world model is cheap and easy to predict, it only captures a single surface of the world. To construct a more complete 4D world model, it may be interesting in the future to have a generative model that generates multiple RGB-DN views of the world, which can then be integrated to form a more complete 4D world model. 

\section*{Acknowledgements}
We are extremely grateful to Pengxiao Han for assistance with the baseline code, and to Yuncong Yang, Sunli Chen, Jiaben Chen, Zeyuan Yang, Zixin Wang, Lixing Fang, and many other friends for their helpful feedback and insightful discussions.

{
    \small
    \bibliographystyle{ieeenat_fullname}
    \bibliography{main}
}
\clearpage
\newpage
   \twocolumn[
    \centering
    \Large
    \textbf{\modelname: Learning 4D Embodied World Models}\\
    \vspace{0.5em}Supplementary Material \\
    \vspace{1.0em}
   ] %
\setcounter{section}{0}
\section{Implementation Details}
\label{app:impl}

\subsection{Video Diffusion Model Details}
We trained an RGB-DN video diffusion model using the CogVideoX~\cite{yang2024cogvideox} architecture. On the input side, our depth normal projector and RGB projector shared the same architecture. On the output side, our \texttt{Conv3DNet} consisted of three layers, while the MLP had two layers, both with a dimension of 1024.
The model outputs videos with 49 frames, utilizing gradient checkpointing to optimize memory usage. We set a global batch size of 16 and used \texttt{bf16} precision to accelerate. For sampling, we employed the DDPM scheduler across 50 steps and set a classifier-free guidance scale of 7.5.

The training spanned 40,000 iterations with an initial learning rate of $1 \times 10^{-4}$, a gradient clipping of 1.0, and a 1,000-step warmup. The optimizer used was Adam with $\epsilon$ set to $1 \times 10^{-15}$, and an exponential moving average (EMA~\cite{klinker2011exponential}) decay of 0.99 was applied to stabilize training.

\subsection{4D Scene Generation}
The parameters for the loss term in Eq.12 are set differently for the RT-1~\cite{brohan2022rt}, Bridge~\cite{walke2023bridgedata} and RLBench~\cite{james2019rlbench} datasets, as shown in the table below. It is worth noting that the selection of \(\lambda\) varies for different scenarios. In practice, achieving the best performance requires tuning these parameters.

\begin{table}[ht]
\centering
\begin{tabular}{ccccc}
\toprule
Dataset  & $\lambda_d$ & $\lambda_b$ & $\lambda_{g1}$ & $\lambda_{g2}$ \\ \midrule
RT-1, Bridge     & 20          & 200         & 20           & 20           \\
RLBench  & 20          & 200         & 2            & 2            \\ \bottomrule
\end{tabular}
\caption{Loss Term Parameters for RT-1 and RLBench Datasets}
\end{table}

\begin{figure}[H]
    \centering
    \includegraphics[width=0.98\linewidth]{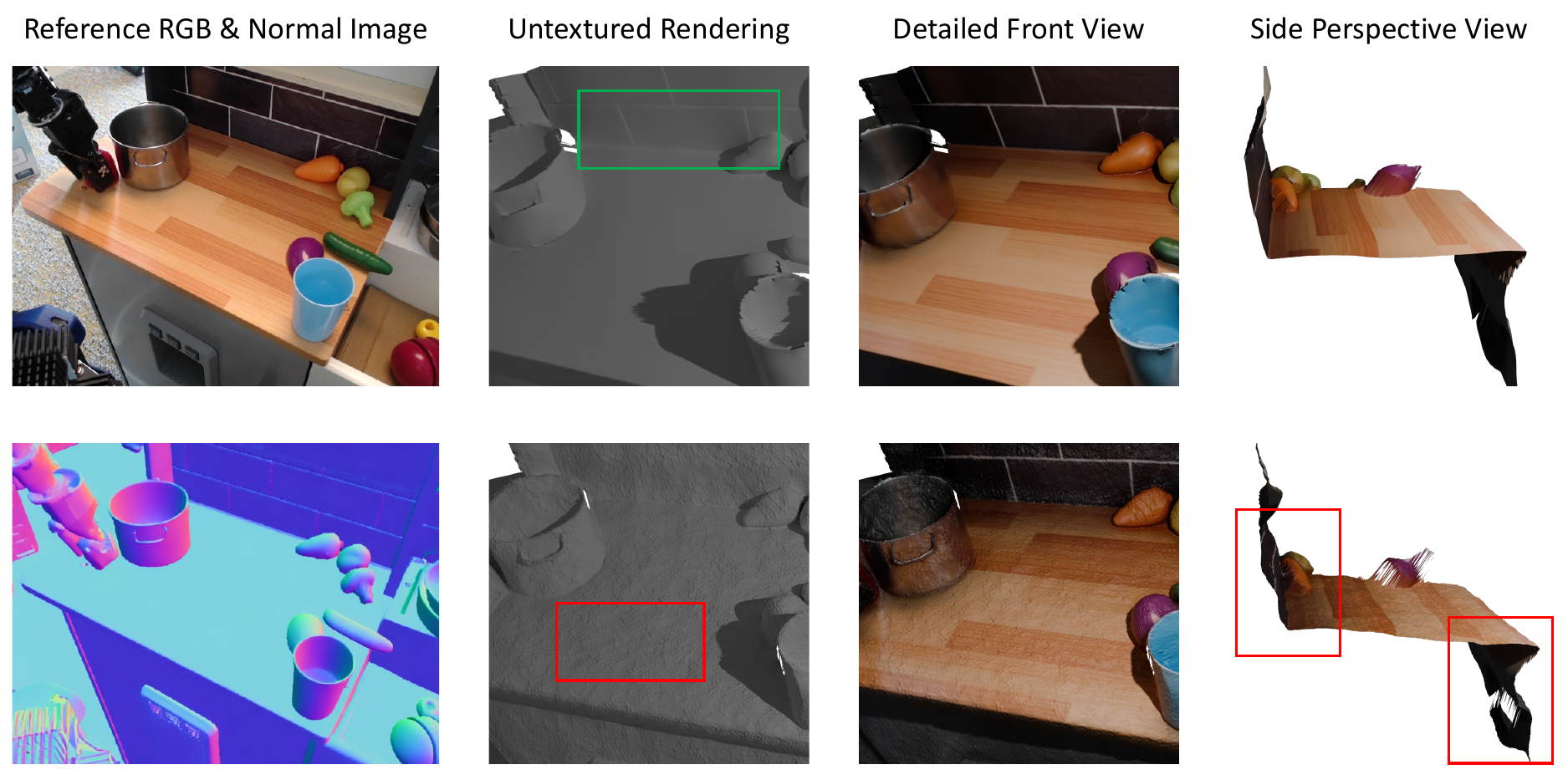}
    \caption{Visualization of the optimized 3D robotic scene reconstruction using our method. The untextured renderings show enhanced detail (green box) and improved surface smoothness (red box). The side perspective view highlights accurate shape and geometry optimization, including the perpendicular alignment of the wall and table (red boxes).}
    \label{fig:bini_works}
    \vspace{-10pt}
\end{figure}

In Figure~\ref{fig:bini_works}, we present a visualization of the 3D robotic scene reconstruction optimized using our proposed method in the BridgeV2~\cite{walke2023bridgedata} dataset. After estimating the depth and normal with the estimator, we refine the outputs to reconstruct the scene accurately. The figure includes untextured rendering and texture-rendered views, where the wall textures are significantly enhanced due to normal optimization. The side perspective view shows the improved shape and geometry reconstruction. Notably, the wall and table surfaces are well-aligned, appearing perpendicular to each other, further validating the effectiveness of our optimization process in capturing accurate spatial relationships.

\begin{figure}[b]
    \centering
    \includegraphics[width=0.98\linewidth]{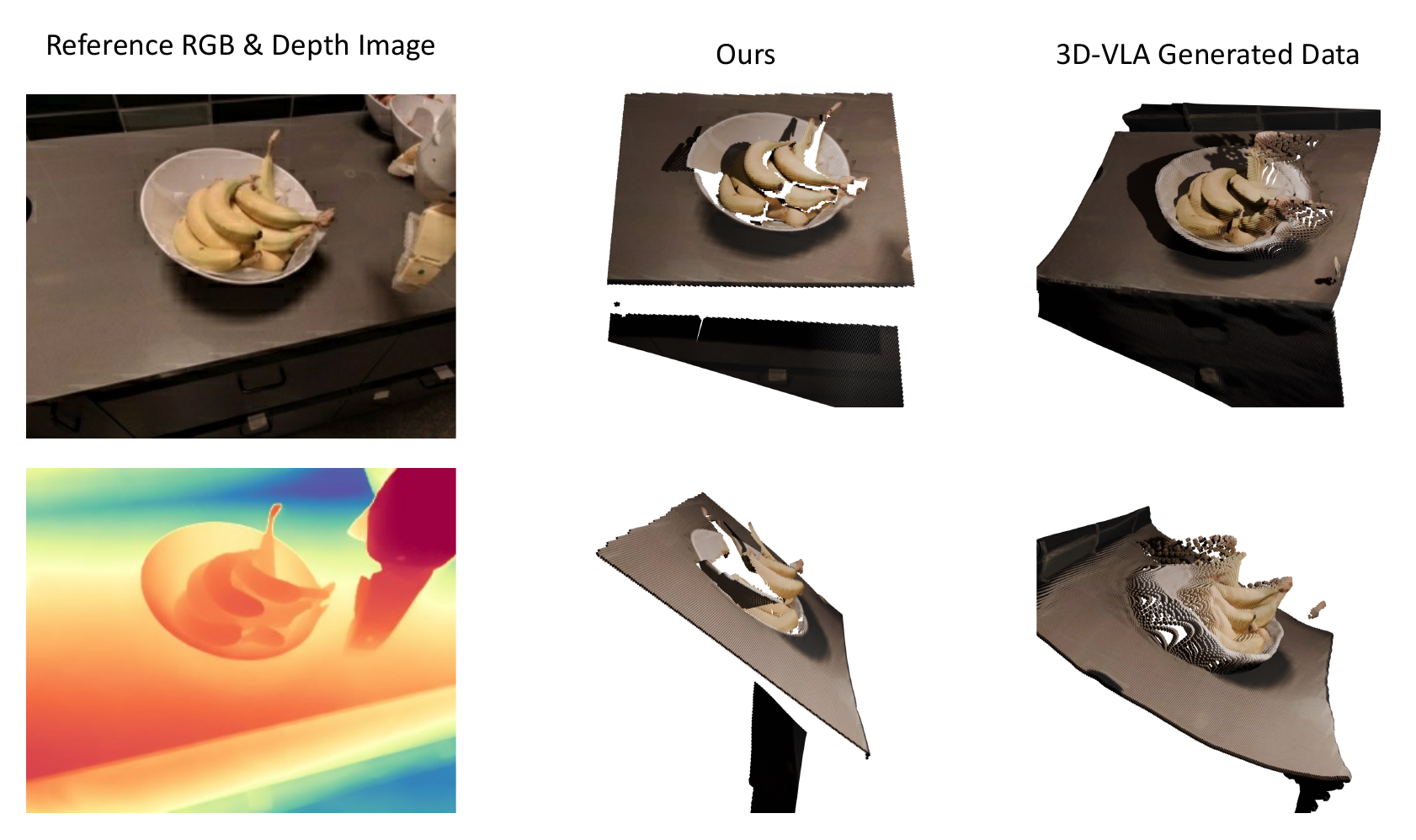}
    \caption{Comparison of point cloud generation quality between our method and 3D-VLA}
    \label{fig:data_exp}
\end{figure}

\subsection{Implementation Details for Robotics Planning}
For the RLBench training, we adopted the same architecture and methods as our video diffusion model, with the primary difference being that we used 13 frames and fine-tuned the model.

For the action prediction stage, we first filter out the background and floor from the data, focusing only on the points of the table and the objects manipulated by the robotic arm, and then sample 8192 points from the filtered point cloud. In our inverse dynamic model, the PointNet extracts features from this point cloud, concatenated with the instruction's language embedding, and passed into a 4-layer MLP, finally outputting the 7DoF actions.
To augment the data and better adapt to the output of video diffusion models, we add significant Gaussian noise (with a relative magnitude of 20\%) to both the image and point cloud coordinates.

\begin{figure*}[t]
    \centering
    \includegraphics[width=0.98\linewidth]{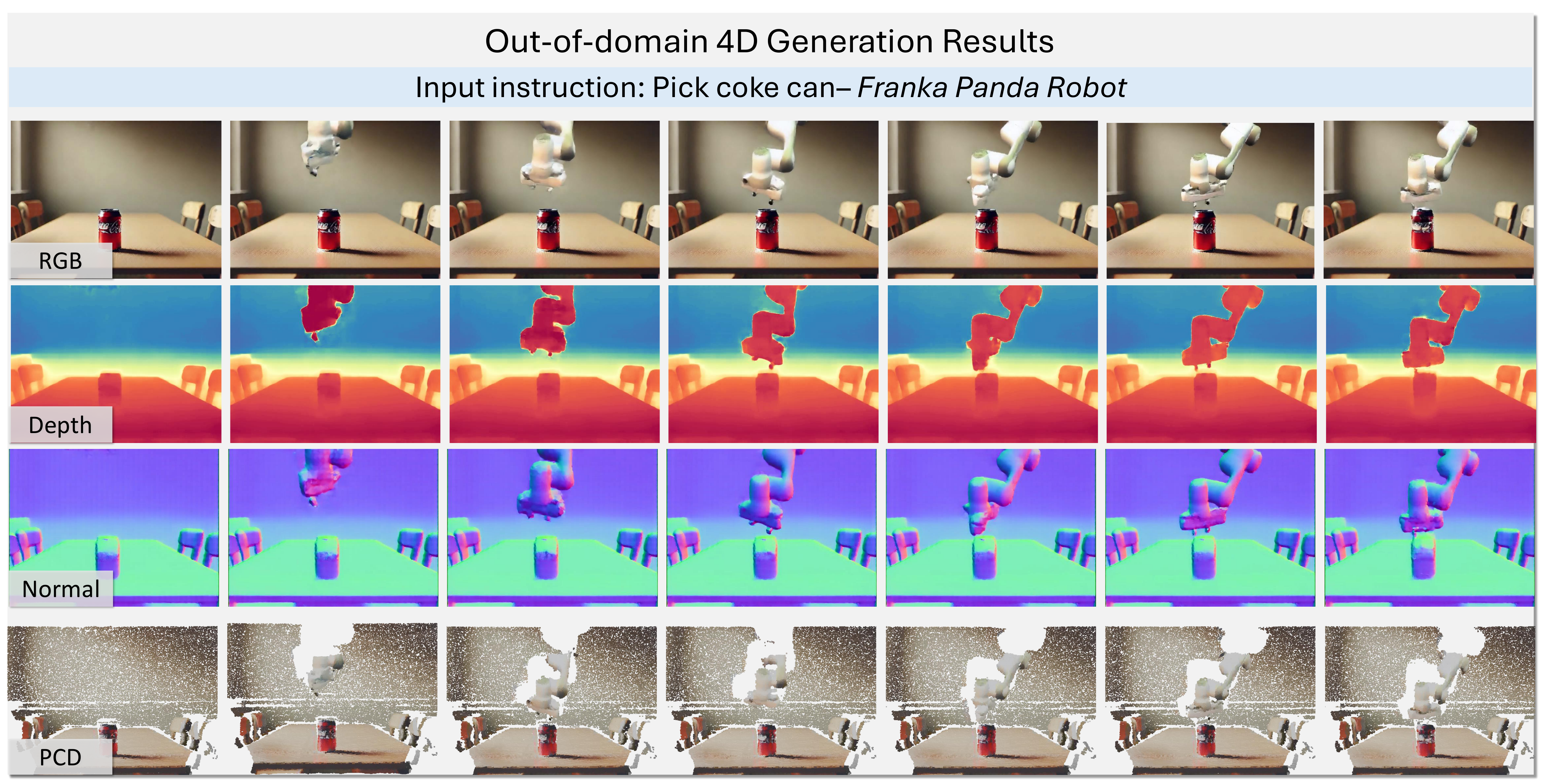}
    \caption{Out-of-domain 4D generation results}
    \label{fig:supp-wild-results}
\end{figure*}

\section{More Qualitative Results}

\subsection{Data Annotation}
\begin{figure*}[t]
    \centering
    \includegraphics[width=0.8\linewidth]{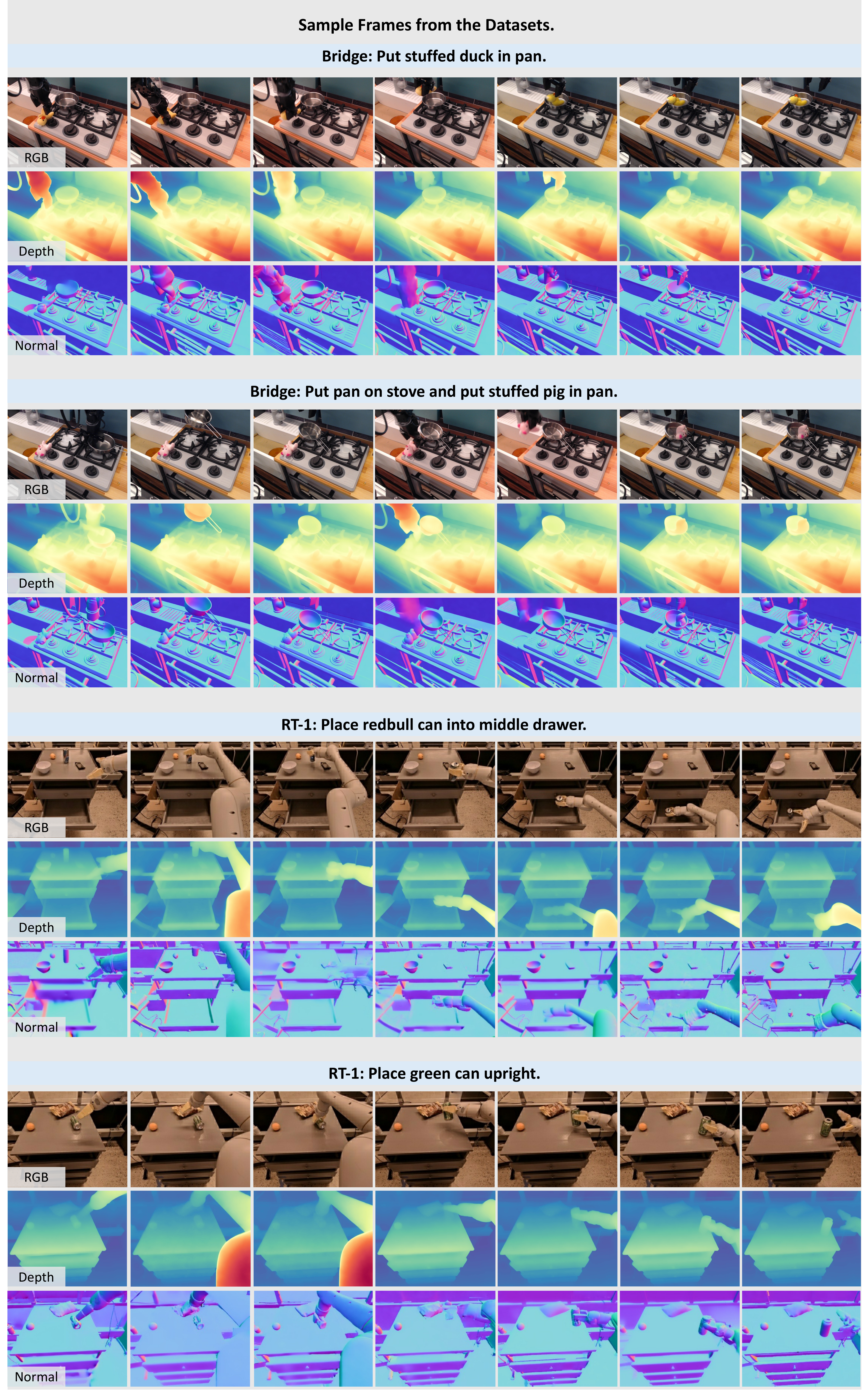}
    \caption{Some sample frames extracted from the datasets Bridge \cite{walke2023bridgedata} and RT-1 \cite{brohan2022rt}. }
    \label{fig:dataset}
\end{figure*}

In this section, we first compare our data generation method with 3D-VLA~\cite{zhen20243d}. They use ZoeDepth~\cite{bhat2023zoedepth} for depth map estimation and directly map them into 3D space. The comparison results, shown in Figure~\ref{fig:data_exp}, evaluate the quality of point cloud generation for both methods, with cubes replacing vertices for rendering. Our generated data demonstrates higher realism, while 3D-VLA exhibits noticeable shape distortion. Figure~\ref{fig:dataset} showcases some of the RGB, depth, and normal images from the datasets we used, along with the corresponding natural language instructions.

\subsection{4D Generation}
As shown in Figure~\ref{fig:supp-wild-results}, we present our out-of-domain results. We used DALL-E~\cite{ramesh2021zero} to generate an image and prompted the 4D world model for generation. Our video diffusion model directly produces the RGB, depth, and normal maps, while the point clouds are rendered from the reconstructed outputs. These results highlight the robustness of our method across different visual modalities.
\textit{Additional video results} can be found in the supplementary materials folder for further analysis and evaluation.

\subsection{Video Generation}
\begin{figure*}[t]
    \centering
    \includegraphics[width=0.98\linewidth]{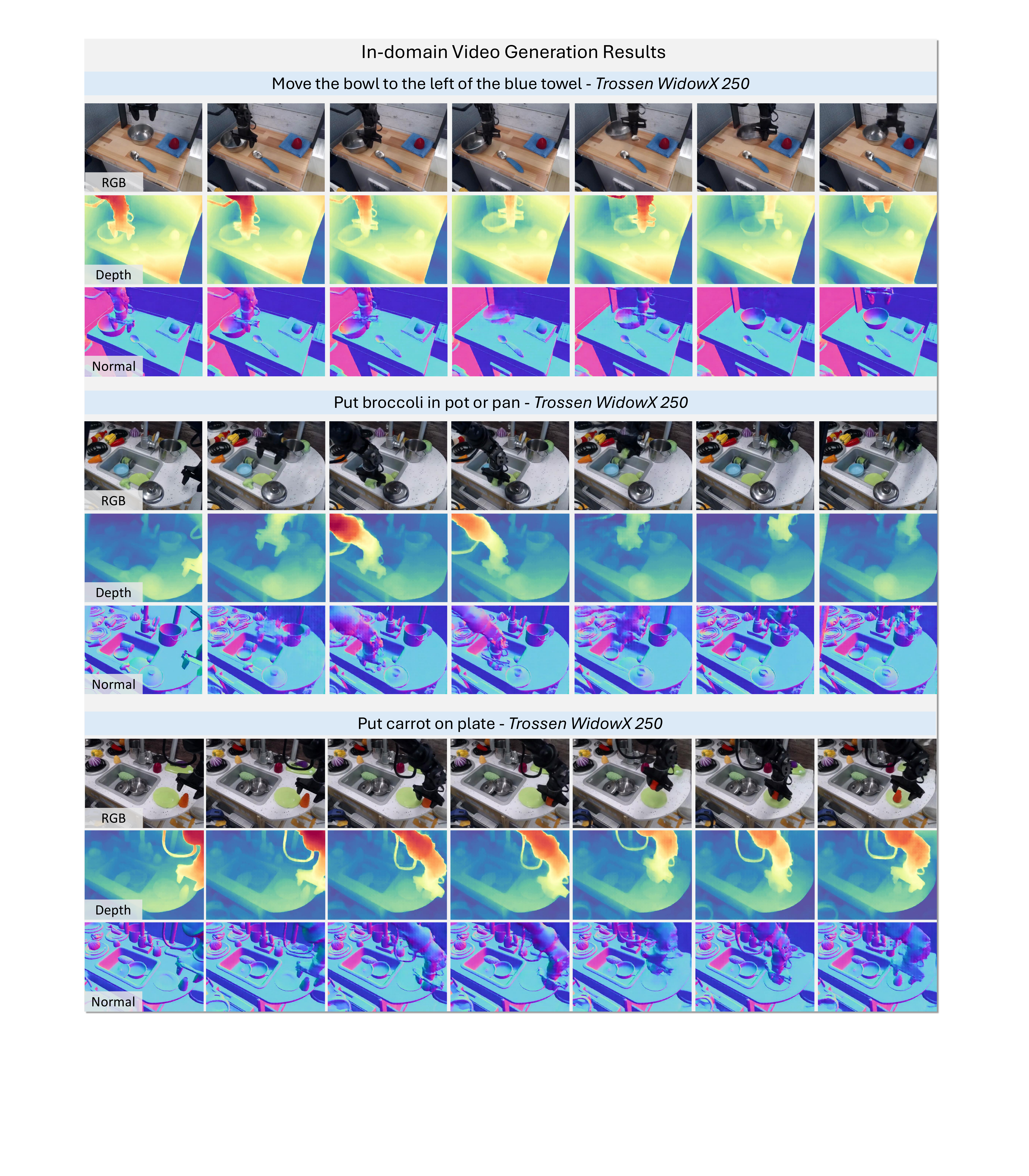}
    \caption{In-domain RGB-DN Video generation results on Bridge dataset}
    \label{fig:supp-bridge-results}
\end{figure*}

\begin{figure*}[t]
    \centering
    \includegraphics[width=0.98\linewidth]{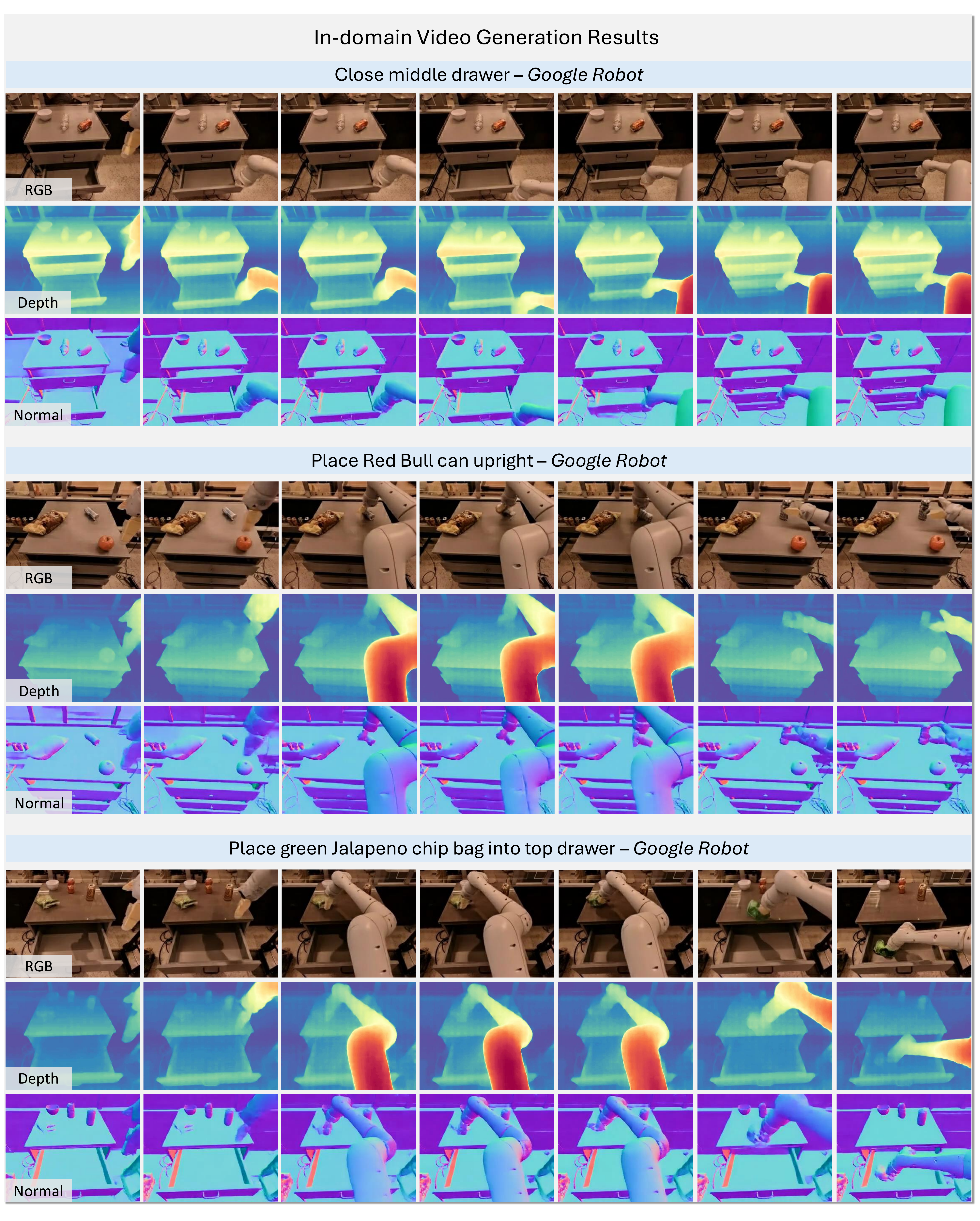}
    \caption{In-domain RGB-DN Video generation results on RT1 dataset}
    \label{fig:supp-rt1-results}
\end{figure*}

\begin{figure*}[t]
    \centering
    \includegraphics[width=0.80\linewidth]{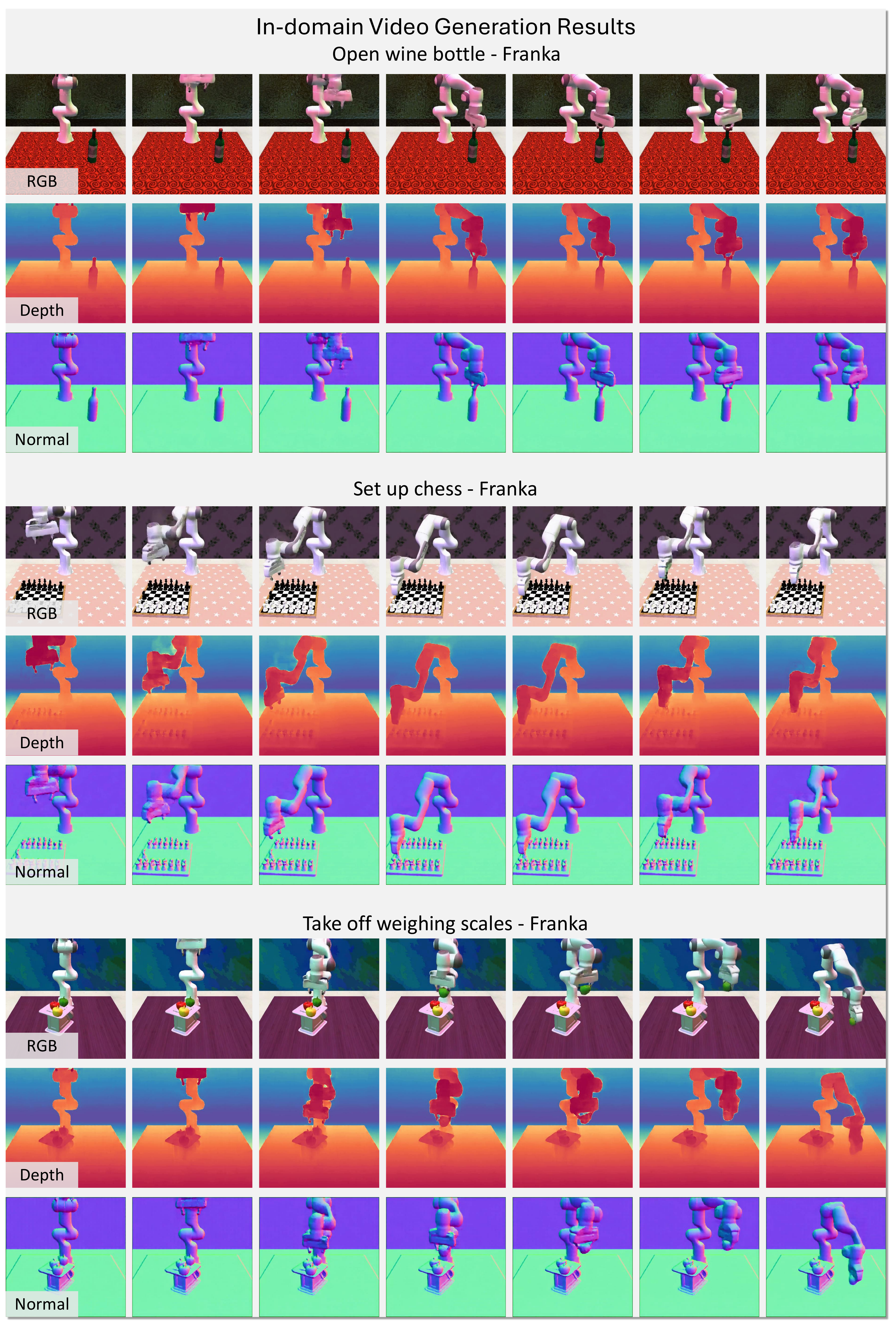}
    \caption{In-domain RGB-DN Video generation results on RLBench dataset}
    \label{fig:supp-rlbench-results}
\end{figure*}

We present the video generation results on the RT1, Bridge and RLBench datasets in Figure~\ref{fig:supp-rlbench-results}, Figure~\ref{fig:supp-bridge-results} and Figure~\ref{fig:supp-rt1-results}. More videos can be found in our supplementary folder.

\subsection{Explicit Action Planning}
\begin{figure}[h]
    \centering
    \includegraphics[width=0.98\linewidth]{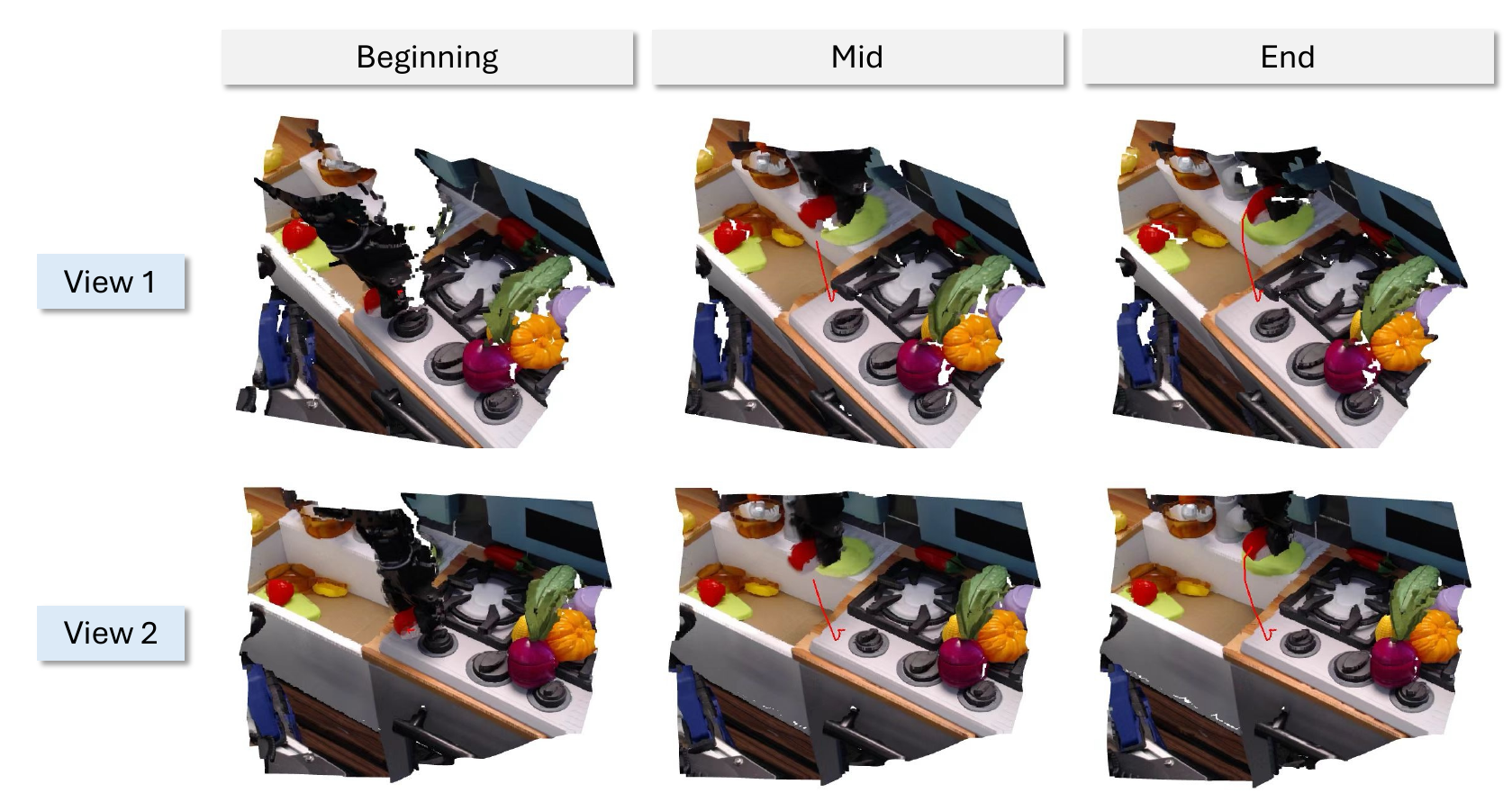}
    \caption{Tracking and visualization of robotic arm action trajectories on the Bridge dataset}
    \label{fig:act_vis}
\end{figure}

One potential application of our generated mesh is to extract action trajectories directly. As illustrated in Figure~\ref{fig:act_vis}, we track the robotic arm in the video to capture its motion path. This trajectory is subsequently lifted into 3D space, enabling the reconstruction of the robot arm's action trajectory. The red line in the visualization represents the captured action trajectory.

\end{document}